\documentclass[conference]{IEEEtran}

\ifCLASSINFOpdf
\else
\fi
\usepackage{url}
\usepackage[]{fancyhdr} %
\newcommand{\changefont}{\fontsize{8}{9}\selectfont} 
\usepackage{url}
\usepackage{verbatim}
\usepackage{graphicx}
\usepackage{color}
\usepackage[table]{xcolor} 
\usepackage{caption}
\usepackage{lipsum}
\usepackage[caption=false]{subfig}
\usepackage{cite}
\usepackage{amssymb}
\usepackage{amsmath}
\usepackage{amsfonts}
\usepackage{placeins}
\usepackage{algorithm}
\usepackage{algorithmicx}
\usepackage{algpseudocode}
\usepackage{multirow}

\algnewcommand{\LeftAlignComment}[1]{\Statex \(\triangleright\) \textit{#1}}
\usepackage{balance}
\usepackage{lipsum}  
\usepackage{mathtools}
\usepackage{tabularx,booktabs}

\usepackage{color}
\usepackage{nomencl}
\makenomenclature
\usepackage{algorithm}
\usepackage{algorithmicx}
\usepackage{algpseudocode}

\def\BibTeX{{\rm B\kern-.05em{\sc i\kern-.025em b}\kern-.08em
    T\kern-.1667em\lower.7ex\hbox{E}\kern-.125emX}}

\usepackage{standalone}
\usepackage{tikz} 
\usepackage{pgfplots} 

\usepackage{dblfloatfix}

\usepackage{float}
\usepackage{adjustbox}
\definecolor{grey}{RGB}{169,169,169}

\IEEEoverridecommandlockouts
\begin{document}

%

\title{Data-driven Models to Anticipate Critical Voltage Events in Power Systems}

\author{\IEEEauthorblockN{Fabrizio De~Caro}
\IEEEauthorblockA{\textit{Dep. of Engineering} \\
\textit{University of Sannio}\\
Benevento, Italy \\
fdecaro@unisannio.it}
\and
\IEEEauthorblockN{Adam J. Collin}
\IEEEauthorblockA{\textit{Dep. of Engineering} \\
\textit{University of Sannio}\\
Benevento, Italy \\
adam.collin@unisannio.it}
\and
\IEEEauthorblockN{Alfredo Vaccaro}
\IEEEauthorblockA{\textit{Dep. of Engineering} \\
\textit{University of Sannio}\\
Benevento, Italy \\
vaccaro@unisannio.it}
}


%





\maketitle
\thispagestyle{fancy}
\pagestyle{fancy}



\begin{abstract}
This paper explores the effectiveness of data-driven models to predict voltage excursion events in power systems using simple categorical labels. By treating  the  prediction  as  a categorical  classification  task, the workflow is characterized by a low computational and data burden. A proof-of-concept case study on a real portion of the Italian 150 kV sub-transmission network, which hosts a significant amount of wind power generation, demonstrates the general validity of the proposal and offers insight into the strengths and weaknesses of several widely utilized prediction models for this application.
\end{abstract}




%
\IEEEpeerreviewmaketitle

\section{Introduction}
\IEEEPARstart{T}{he} ongoing integration of renewable generation has many, well-reported, impacts on power system operation and control. One area of  concern for network operators is the variations in voltage magnitude brought about by the increasing amount of wind and solar generation connected to sub-transmission and distribution networks. As voltages must be regulated within a prescribed range to ensure secure and effective network operation, avoiding voltage excursions, i.e., deviations from the operating range, is of utmost importance \textcolor{black}{and requires system operators to adopt new network management controls.}

\textcolor{black}{Accordingly}, the prediction of the voltage magnitude, \textcolor{black}{not historically} widely considered as a standalone activity, is receiving more interest as a research activity \cite{voltage_predict1, voltage_predict2, voltage_predict3}. Traditionally, voltage magnitude is considered as part of state estimation (SE) tools. However, SE requires large amounts of data, from multiple sources, and must account for uncertainties in network configuration, generation/load variability and node observability~\cite{review_se}. \textcolor{black}{Due to the large volume of data involved, machine learning (ML) techniques} offer opportunities in this application area \cite{de2020review}.

\textcolor{black}{ML techniques have found widespread use in power system event detection and prediction tasks, e.g., load forecasting \cite{ml_load_forecast}, fault classification and prediction \cite{ml_fault} and reliability and security analysis \cite{ml_reliability_review}. In the context of voltage event detection and control, several different ML techniques have been proposed, including: online voltage security assessment of high voltage networks using phasor measurment unit data and decision trees \cite{dt_pmu} and deep reinforcement learning to update control settings in real time, as opposed to traditional offline analysis \cite{drl}. Another area of research, of direct relevance to this paper, is the prediction of the voltage itself. This can be considered important as forecasting the voltage allows the system operator additional time to prepare for the implementation of possible control actions. This has been considered for the prediction of voltage magnitudes in low voltage networks using a deep learning neural network architecture \cite{dlnn}, however, this technique relies on knowledge of the network impedance (i.e. configuration). A purely data-driven approach was proposed in \cite{voltage_predict3}, using various single and ensemble regressors to predict the voltage magnitude in distribution networks.}

In this paper, the network voltage is considered in isolation and the prediction problem is treated as a categorical classification task, rather than a regression problem. Accordingly, the data requirements, and associated uncertainty, are significantly reduced compared to other approaches. This affords the possibility of (accurately) predicting over longer horizons with a low computational burden, and such a method can be used to anticipate events and improve the situational awareness of the network operator~\cite{situation_aw}. Real data obtained from a portion of the  Italian  150~kV  sub-transmission  network, which hosts a significant amount of wind power generation, is used in a proof-of-concept study, which includes the appraisal of different analysis metrics and data-driven prediction models for this new application. \textcolor{black}{It is well known that different ML metrics are suitable for different types of data/classification tasks. As such, the appraisal of different ML metrics is considered an important aspect of the paper as their use in evaluating the proposed categorical classification task provides insight into the performance of the model itself. By considering the relationship between the data and the ML metrics, the paper also identifies the most effective metric for evaluating this type of classification task.}








\section{Voltage Excursion Event Predictor}
\begin{figure}[!t]
    \centering
    \includegraphics[width=0.9\columnwidth,keepaspectratio,clip,trim={0cm 22.0cm 10.9cm  1cm}]{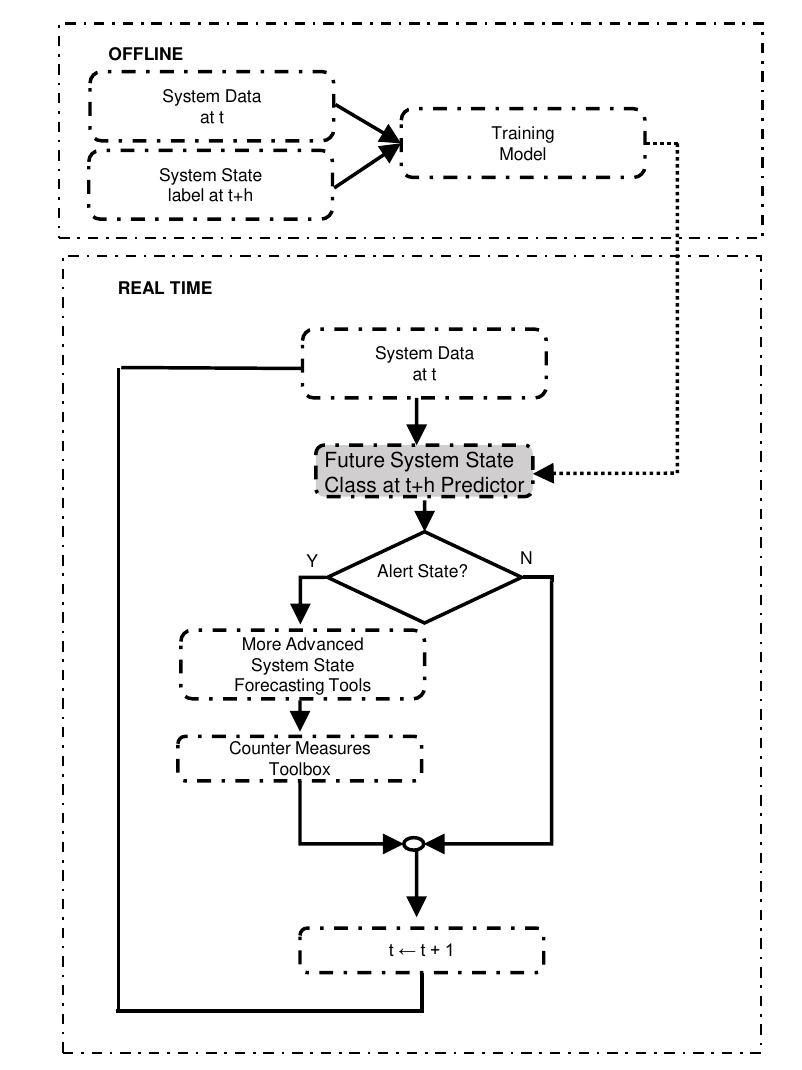}
    \caption{Workflow of the voltage excursion event predictor tool.}
    \label{fig:scheme}
\end{figure}

Voltage excursion event forecasting consists of predicting voltage magnitude values close to, or out of, the voltage tolerance bounds. The conceptualized framework shown in Fig.~\ref{fig:scheme} anticipates possible critical voltage events of the grid ahead of real-time, through a hierarchical-based approach. Given the generic time sample $t$, the grey-coloured block, which represents the predictor model, returns the label of the possible state of the grid $t+h$ from the available data acquired from the energy management system. If the predicted label matches a critical event for the grid, more detailed simulations are conducted to assess, and implement, adequate countermeasures. 

\textcolor{black}{
The more detailed simulations depend on the type of predicted critical events, but they can include static (e.g., Optimal Power Flow) and dynamic simulations. Toolboxes aimed at predicting the critical events are trained with historical data about the system operation.}


\subsection{Mathematical Preliminaries}

Given an upper and lower bound for the voltage $\underline{v}$ and $\overline{v}$, and the voltage amplitude time series $\mathcal{X}_{b}=\{x_{b,1},\dots,x_{b,t}\}$ measured at the generic $b$-th bus, 
it is possible to define the over-bound excursion event time series $\mathcal{Y}_b^{(ob)}$ and the under-bound excursion event $\mathcal{Y}_b^{(ub)}$ using:

\begin{equation}
   y_{b,t}^{(ob)} = \begin{cases}
    1 & \textrm{if} \ x_{b,t} > \overline{v} \\
    0 & \textrm{elsewhere}
    \end{cases}
\end{equation}
\begin{equation}
   y_{b,t}^{(ub)} = \begin{cases}
    1 & \textrm{if} \ x_{b,t} < \underline{v} \\
    0 & \textrm{elsewhere}
    \end{cases}
\end{equation}

\noindent where $y_{b,t}^{(ob/ub)} = 1/0$ are defined as a positive/negative events, respectively. The mathematical definitions over- and under-bound are used in place of the power system terms over- and undervoltage as the setting of $\overline{v}$ and $\underline{v}$ is arbitrary, e.g., the network operator may wish to set a threshold lower/greater than an over/undervoltage event.

The  voltage excursion event predictor model $f$ predicts the voltage excursion labels $h$ time steps ahead as follows:

\begin{equation}
\label{eq:MIMO}
    \begin{array}{l}
     y^{(ob/ub)}_{1,t+h} \\
         \dots  \\ 
      y^{(ob/ub)}_{B,t+h}
      \end{array} = f \Bigg( \begin{array}{cc}
    x_{1,t-d-L},\dots,x_{1,t-d}  \\
      \dots  \\ 
    x_{B,t-d-L},\dots,x_{B,t-d}
\end{array} \Bigg)
\end{equation}

\noindent where $d$ is the delay, $L$ is the lag value, $h$ is the forecast horizon and $B$ is the number of buses.


The Multi-Input Multi-Output model described by (\ref{eq:MIMO}) may be difficult to train, hence it is proposed to decompose the problem into many sub-problems, where each local model returns the label prediction for each $b$-th bus. 

This task can be achieved using categorical predictor models, which are classification models where the label linked to each input at $t$ is related to something in the future at $t+h$. Several categorical predictor models are available but the general functionality can be represented by the function $f$ returning the probability score of a certain event given a certain input:


\begin{equation}
\label{eq:generic_f}
    \mathcal{P}(\hat{y}^{(ob/ub)}_{b,t+h} = c | \mathcal{X}_{b,t-L}) \leftarrow f(\mathcal{X}_{b,t-L})\quad \in [0,1]
\end{equation}

\noindent where $c$ is the event label ($0$ or $1$ in this case).

Hence, to transform a probability score into the predicted label, a threshold $\beta$ must be set:

\begin{equation}
\label{eq:generic_beta}
\hat{y}^{(ob/ub)}_{b,t+h} = \begin{cases} 1  & \textrm{if}\ \mathcal{P}^{(ob/ub)}_{t+h} \geq \beta
 \\ 0 & \textrm{elsewhere}
    \end{cases}
\end{equation}

\noindent where $\beta\ \in [0,1]$.

The role of $\beta$ is crucial as it affects the performance of the label predictor models. Low values of $\beta$ produce too many false alarms; conversely, if the value of $\beta$ is too high then positive events may be missed. This is discussed further in following subsection.

\subsection{Comparison Metrics and Visual Inspection Analysis}

The performance of the predictor models is assessed through a set of metrics, which are computed starting from the confusion matrix \cite{tharwat2020classification}: 

\begin{table}[h]
\centering
\caption{Confusion Matrix}
\begin{tabular}{ll|ll}
                  &  & \multicolumn{2}{c}{$\hat{y}^{(ob/ub)}_{b,t+h}$} \\  
                  &  &      1     &     0     \\ \hline
\multirow{2}{*}{$y^{(ob/ub)}_{b,t+h}$} & 1 &   TP    & FN          \\ 
                  & 0 &  FP     &  TN  \\       \hline
\end{tabular}
    \label{tab:my_label}
\end{table}

\noindent where TP, FN, TN, and FP are the number of ``True Positive’’, ``False Negative’’, ``True Negative’’, and ``False Positive’’, respectively, observed in the validation set.

These instances compare the predicted classes with the actual ones. A TP instance is when the positive event is correctly predicted by the model, whereas an FN instance is when the model does not predict the positive event. A TN instance is when the positive event does not occur and the model does not predict it. An FP instance is when the model predicts a positive event that will not occur.

A first metric to consider the performance of a categorical predictor model is to take into account the accuracy (ACC) of the model, defined as the ratio of the correctly predicted instances to all instances:

\begin{equation}
    \textrm{ACC} = \dfrac{TP+TN}{TP+FN+TN+FP} \in [0,1]
\end{equation}

Although this metric is useful in balanced data sets (where the number of positive and negative labels is comparable), in the presence of unbalanced data sets it may be not reliable. For this reason, it is pertinent to assess the performance of the model considering the following ratios:

\begin{equation}
 TPR =  \dfrac{TP}{TP+FN}\ \in [0,1]
\end{equation}

\begin{equation}
 FPR = \dfrac{FP}{TN+FP}\ \in [0,1]
\end{equation}

\begin{equation}
 TNR = \dfrac{TN}{TN+FP}\ \in [0,1]
\end{equation}

\begin{equation}
 FNR = \dfrac{FN}{FN+TP}\ \in [0,1]
\end{equation}




\noindent where TPR, FPR, TNR, and FNR are the True Positive Ratio, the False Positive Ratio, the True Negative Ratio and the False Negative Ratio, respectively. 

However, as explained in (\ref{eq:generic_f}), the vast majority of prediction models return the posterior probability values about each class given an input and a threshold $\beta$ must be set to link the probability to a label to return the categorical prediction (\ref{eq:generic_beta}).


\begin{figure}
\centering
 \includegraphics[width=0.8\linewidth,keepaspectratio]{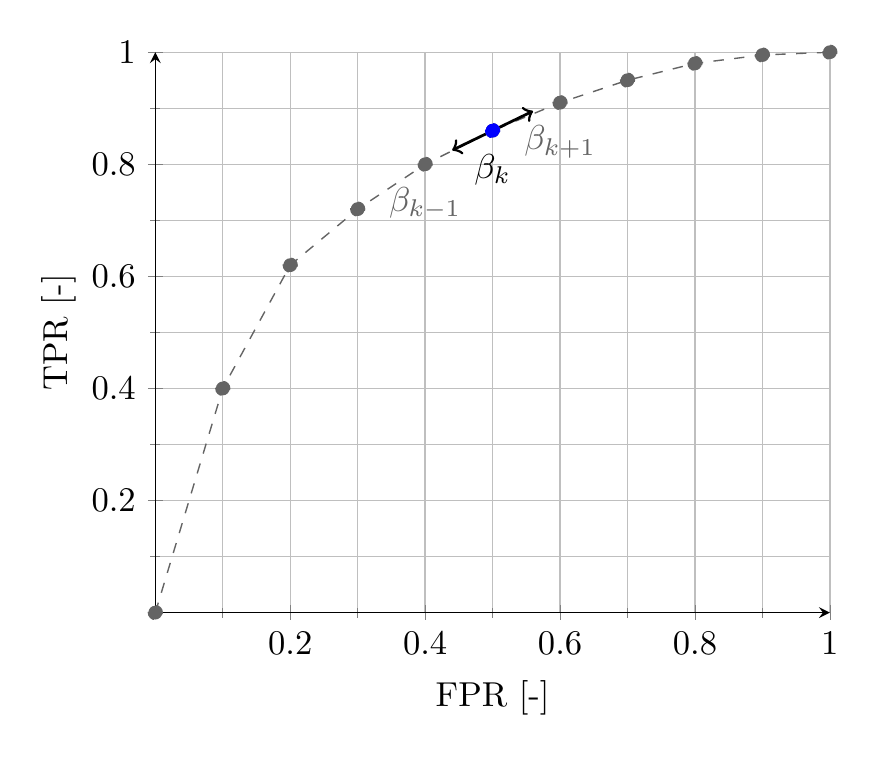}
   \caption{The ROC curve and the $\beta_k=\beta^{*}$ trade-off study, where different FPR-TPR pairs are obtained by varying $\beta$~from~[0,1].}
    \label{fig:ROC}
\end{figure}

The impact of this threshold is typically quantified by analysing the receiver operating characteristic (ROC) curve, which displays the TPR against the FPR. 
The ROC curve is built by computing TPR and FPR for different values of $\beta$ between $0$ and $1$ \cite{bradley1997use}. This process is illustrated in Fig.~\ref{fig:ROC}.

The area under the curve (AUC) metric is the area under the ROC curve and can take any value within the range $[0,1]$. The greater the value of AUC, the better the performance of the model. If $\textrm{AUC} \leq 0.5$, the performance of the model is worse than a random classifier and cannot be considered acceptable as a classification model~\cite{tharwat2020classification}.

However, AUC values cannot supply any information about the trade-off point between TPR and FPR of the prediction model. To extract the trade-off point, the $\beta^*$ value maximizing the Geometric Mean or ``G-Means'' metric (\ref{eq:geometric_mean}) is identified.

\begin{equation}
\label{eq:geometric_mean}
    GM = TPR \cdot (1 - FPR) = TPR \cdot TNR\ \in [0,1]
\end{equation}

$GM$ returns information about the prediction balance between TPR and FPR. The highest value of $GM$ returns the operative point maximizing this balance. The lower the maximum $GM$ value is, the poorer the positive event prediction performance is, even if negative events are correctly predicted.



\begin{figure*}[!b]
    \centering
    \includegraphics[width=16cm,keepaspectratio,clip,trim={0cm 0cm 0cm 1.25cm}]{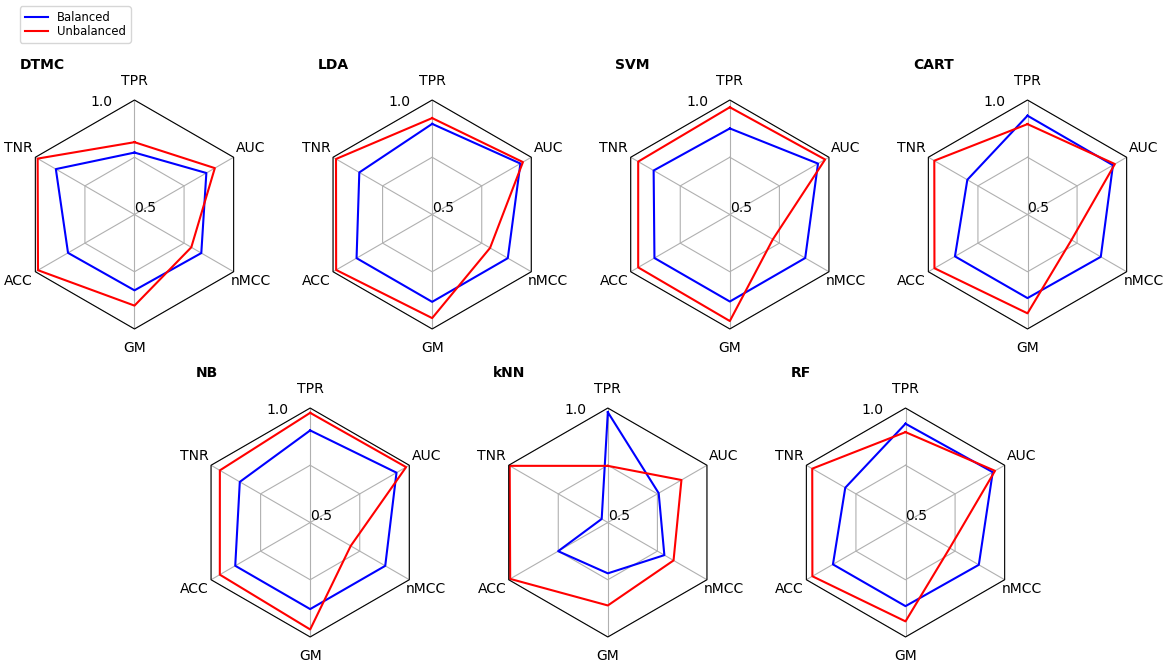}
    \caption{Performance of the predictor models classified by several metrics. The results for the balanced data set are shown in black, with results for the unbalanced data set in red. The closer the points are to the vertex, the more effective the model is for the stated metric.}
    \label{fig:radar}
\end{figure*}

The final means of comparison considered in this paper is the Matthews Correlation Coefficient (MCC) metric:

\begin{equation}
MCC = \dfrac{TP \cdot TN-FP\cdot FN}{\sqrt{ \splitfrac{(TP+FP)\cdot(TP+FN)\cdot}{(TN+FP)\cdot(TN+FN)} }} \in [-1,1] 
\end{equation}

MCC is the only metric which returns a high score only if the model correctly predicts both positive and negative events~\cite{chicco2020advantages}. To recast the range of MCC to that of the other metrics, the nMCC is computed as:

\begin{equation}
nMCC = \dfrac{MCC+1}{2} \in [0,1] 
\end{equation}

\noindent where $nMCC<0.5$ corresponds to using a random classifier.

As shown, several metrics may be utilized to compare the performance of different predictors and selecting the appropriate criterion requires careful consideration \cite{strbac_dt}. However, the use of nMCC as a compact means of comparison is advocated in this research and discussed further in the Case Study.

\section{Case Study}

\subsection{Experimental Setup}
The voltage excursion event predictors models are analyzed using real network data from a portion of the Italian $150$~kV sub-transmission network consisting of 15 buses. The network hosts a significant amount of wind power generation and, as such, is susceptible to voltage rise. The case study data consists of eight months of voltage amplitude sampled every $15$ minutes. The $28,934$ samples for each bus were split into training and test sets according to a $3$:$1$ ratio. The forecasting horizon was set to $1$ hour, i.e., $h=4$. The considered lag was equal to one ($L=1$) with zero delay $d = 0$, i.e., each model predicts the event at $t+h$ using the last two samples at $t$ and $t-1$. The models were tested for two thresholds: $\overline{v}=1.05\ p.u$ and $\overline{v}=1.08\ p.u$. In this way the performance of different models (\ref{eq:generic_f}) can be assessed for different frequencies of positive event occurrence, as indicated in Table~\ref{tab:all_bus_events}.

\begin{table}[!h]
\caption{Ratio of positive and negative events for two voltage thresholds. The buses selected as balanced and unbalanced data sets are indicated in grey.}
\centering
\label{tab:all_bus_events}
\resizebox{0.8\columnwidth}{!}{%
\begin{tabular}{@{}lrrrr@{}}
\toprule[1pt]
\toprule[0.3pt]
Bus N   & \multicolumn{2}{c}{$\overline{v} \geq 1.05$}  & \multicolumn{2}{c}{$\overline{v} \geq 1.08$   }  \\
        & Pos (\%)  & Neg. (\%) & Pos. (\%) & Neg. (\%) \\ \midrule
1	&	7.6	&	92.4	&	0.3	&	99.7	\\
2	&	7.6	&	92.4	&	0.3	&	99.7	\\
3	&	10.1	&	89.9	&	0.4	&	99.6	\\
4	&	39.1	&	60.9	&	4.9	&	95.1	\\
5	&	61.3	&	38.7	&	13.1	&	86.9	\\
\cellcolor{gray!25}{\textbf{6}}	&	15.8	&	84.2	&	\cellcolor{gray!25}{\textbf{1.0}}&	\cellcolor{gray!25}{\textbf{99.0}}	\\
7	&	11.2	&	88.8	&	0.3	&	99.7	\\
8	&	31.8	&	68.2	&	3.3	&	96.7	\\
9	&	48.2	&	51.8	&	7.4	&	92.6	\\
10	&	58.3	&	41.7	&	11.8	&	88.2	\\
11	&	52.5	&	47.5	&	8.6	&	91.4	\\
\cellcolor{gray!25}{\textbf{12}}&	\cellcolor{gray!25}{\textbf{48.7}}	&	\cellcolor{gray!25}{\textbf{51.3}}	&	7.6	&	92.4	\\
13	&	32.0	&	68.0	&	3.4	&	96.6	\\
14	&	53.5	&	46.5	&	9.7	&	90.3	\\
15	&	53.0	&	47.0	&	9.4	&	90.6	\\ \bottomrule[0.3pt]
\bottomrule[1pt]
\end{tabular}%
}
\end{table}


\subsection{Predictor Models Description}
\label{sec:case_study_models}
Several different machine learning (ML) models and statistical methods may be applied for the task of categorical prediction. Due to space limitations, and to retain the focus on the power system application, the models considered in this research are simply listed here alongside detailed references on their functionality:

\begin{itemize}
    \item Classification And Regression Tree (CART) \cite{lewis2000introduction} 
    \item Random Forest (RF) \cite{breiman2001random}
    \item k-Nearest Neighbors (kNN) \cite{fix1989discriminatory} 
    \item Support Vector Machine (SVM) \cite{corinna1995vapnik} 
    \item Naive-Bayes (NB) \cite{friedman1997bayesian} 
    \item Linear Discriminant Analysis (LDA) \cite{balakrishnama1998linear} 
    \item Discrete Time Markov Chain (DTMC) \cite{carpinone2015markov}
\end{itemize}

As the analysis considers $L=1$, with two input variables per sample for the ML-based models, a $2$\textsuperscript{nd} order DTMC was utilized for a fair comparison.

\subsection{Experimental Results}

The experimental results are analyzed in detail for the cases indicated in Table \ref{tab:all_bus_events}, with summarizing results reported for the other buses. 
The performance of each model was assessed at its trade-off point between TPR and FPR, which is the $\beta^{*}$ value maximizing the GM score for each model. The value of $\beta^{*}$ was identified using an iterative procedure.

\subsubsection{Balanced Data Set}

Bus 12 with $\overline{v}$ equal to $1.05$ is the most balanced data set (Table \ref{tab:all_bus_events}). A visual comparison of all metrics is shown in Fig.~\ref{fig:radar} using the black curves, with data values included in Table~\ref{tab:balanced}. The radar plots demonstrate the difficultly in identifying the best performing model by visual methods. However, an inspection of Fig.~\ref{fig:radar} reveals that the majority of the models have comparable performance in the case of balanced data sets. The noticeable exceptions are DTMC and kNN. The data in Table \ref{tab:balanced} confirms that DTMC predicts too many FN events, with kNN predicting too many FP events. 

\subsubsection{Unbalanced Data Set}
If the $\overline{v}$ is increased to $1.08$ the number of positive events significantly reduces (Table \ref{tab:all_bus_events}). For Bus 6, only 1\% of the samples are positive samples. In this case, the ACC of all models is very high but it is skewed by the high number of negative events. Although DTMC and kNN have the joint highest ACC score, and DTMC has a higher GM score, Table~\ref{tab:bus6_v1.05} reveals that kNN performs significantly better with regards FP and TN, which are only fully considered in nMCC. This test case demonstrates the value of the nMCC metric and reveals that kNN is the best model for event prediction in this scenario.



\subsubsection{Trade-off Point}

Table~\ref{tab:balanced} and~\ref{tab:bus6_v1.05} show that the balance of the data set can affect the trade-off point $\beta^{*}$ for the majority of the models. CART, LDA, kNN, and NB models exhibit the smallest variation. This aspect should not be neglected, as it indicates whether a model requires tuning, i.e., a unique $\beta^{*}$, for each bus and voltage threshold, which translates into increased model complexity from the network operator's perspective.




\begin{table}[!h]
\centering
\caption{Detailed results of the balanced data set}
\label{tab:balanced}
\resizebox{0.89\columnwidth}{!}{%
\begin{tabular}{@{}lrrrrrrr@{}}
\toprule[1pt]
\toprule[0.3pt]
	&	DTMC	&	CART	&	LDA	&	SVM	&	NB	&	kNN	&	RF	\\ \midrule
$\beta^{*}$	&	0.25	&	0.65	&	0.65	&	0.73	&	0.68	&	0.53	&	0.63	\\
TP	&	7237	&	8750	&	8414	&	8227	&	8474	&	9221	&	8749	\\
FP	&	1023	&	1945	&	1312	&	1145	&	1436	&	4639	&	1941	\\
FN	&	2156	&	643	&	979	&	1166	&	919	&	172	&	644	\\
TN	&	8869	&	7947	&	8580	&	8747	&	8456	&	5253	&	7951	\\
TPR	&	0.77	&	0.93	&	0.90	&	0.88	&	0.90	&	0.98	&	0.93	\\
FPR	&	0.10	&	0.20	&	0.13	&	0.12	&	0.15	&	0.47	&	0.20	\\
GM	&	0.83	&	0.87	&	0.88	&	0.88	&	0.88	&	0.72	&	0.87	\\
ACC	&	0.84	&	0.87	&	0.88	&	0.88	&	0.88	&	0.75	&	0.87	\\
nMCC	&	0.84	&	0.87	&	0.88	&	0.88	&	0.88	&	0.78	&	0.87	\\

\bottomrule[0.3pt]
    \bottomrule[1pt]
\end{tabular}%
}
\end{table}

\begin{table}[!h]
\centering
\caption{Detailed results of the unbalanced data set}
\label{tab:bus6_v1.05}
\resizebox{0.89\columnwidth}{!}{%
\begin{tabular}{@{}lrrrrrrr@{}}
\toprule[1pt]
\toprule[0.3pt]
	&	DTMC	&	CART	&	LDA	&	SVM	&	NB	&	kNN	&	RF	\\ \midrule
$\beta^{*}$	&	0.03	&	0.53	&	0.53	&	0.03	&	0.53	&	0.53	&	0.25	\\
TP	&	155	&	170	&	175	&	184	&	186	&	142	&	170	\\
FP	&	223	&	573	&	289	&	724	&	852	&	97	&	566	\\
FN	&	35	&	20	&	15	&	6	&	4	&	48	&	20	\\
TN	&	18872	&	18522	&	18806	&	18371	&	18243	&	18998	&	18529	\\
TPR	&	0.82	&	0.89	&	0.92	&	0.97	&	0.98	&	0.75	&	0.89	\\
FPR	&	0.01	&	0.03	&	0.02	&	0.04	&	0.04	&	0.01	&	0.03	\\
GM	&	0.90	&	0.93	&	0.95	&	0.97	&	0.97	&	0.86	&	0.93	\\
ACC	&	0.99	&	0.97	&	0.98	&	0.96	&	0.96	&	0.99	&	0.97	\\
nMCC	&	0.79	&	0.72	&	0.79	&	0.72	&	0.70	&	0.83	&	0.72	\\

\bottomrule[0.3pt]
    \bottomrule[1pt]
\end{tabular}%
}
\end{table}

\subsubsection{Network Area Results}
\balance

\begin{figure}[h!]
    \centering
    \includegraphics[width=0.95\linewidth,keepaspectratio]{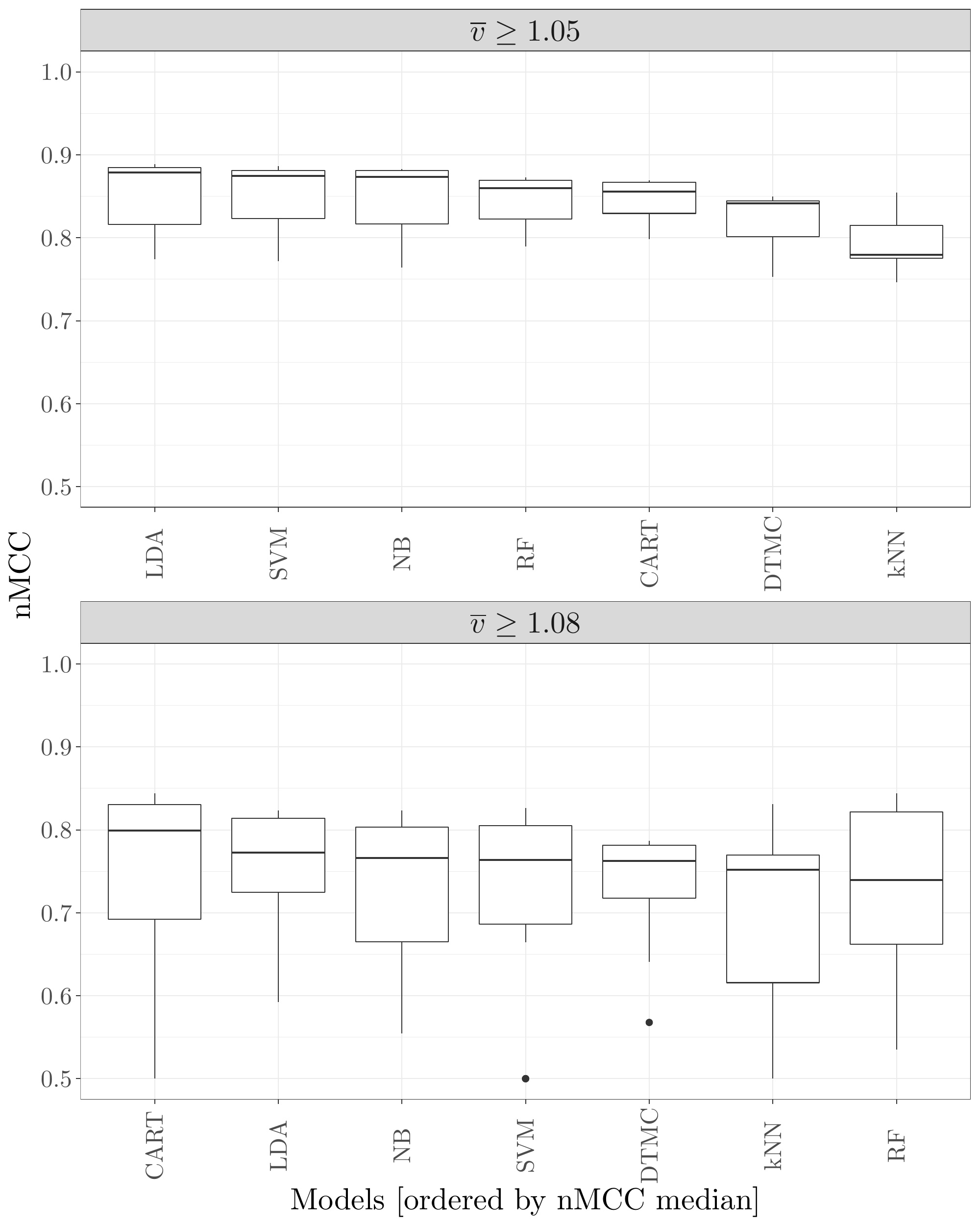}
    \caption{nMCC boxplots considering all buses of the grid. The models are arranged in descending order considering the median value of nMCC calculated from all buses.}
    \label{fig:boxplot}
\end{figure}

As it has been shown that the nMCC is an efficient metric to compare different event predictor models, without reverting to visual analysis, the other metrics are not considered further in this analysis. Fig.~\ref{fig:boxplot} shows the distribution of the nMCC values obtained using data from all 15 buses for the two given voltage thresholds. The models are ranked with respect to the median value of nMCC calculated across all buses for each voltage threshold.

For the case of $\overline{v} \geq 1.05$, the performance is fairly uniform over all system buses. However, RF and CART merit special attention because their nMCC spread is the lowest, i.e., the models have similar performance over all buses.

Given $\overline{v} \geq 1.08$, the prediction task becomes harder because the number of positive events in the training set decreases, reducing the training capacity of the models. Hence, a general deterioration of performance is observed and the range in nMCC values increases for all models. DTMC and SVM have the lowest spread of the model set, and this is a positive factor to consider. The RF and CART models, which had the best performance when $\overline{v} \geq 1.05$, return a wide range of nMCC values in this case, with values approaching 0.5, i.e., equivalent to a random classifier. Despite having the best performance in the unbalanced data set scenario, when considered across the network area, the kNN model exhibits undesirable characteristics in both cases: lowest median when $\overline{v} \geq 1.05$ and wide range when $\overline{v} \geq 1.08$. 


\section{Conclusion}

Situational awareness in complex systems relies not only on the visibility of the existing state of the system, but also on the capability to predict the future state evolution, to detect possible critical states ahead of real-time, and to alert the system operator to investigate them in order to promptly implement effective countermeasures.

In this context, this paper proposed a possible workflow of an alert system for the prediction of voltage excursion events. By treating the prediction as a categorical classification task, rather than a regression  problem, the workflow is characterized by a low computational and data burden. The paper has compared the performance of different system event predictor models from two perspectives: firstly, to find effective metrics to assess these models without having to revert to complicated visual inspection analysis, and secondly, to assess the performance of these models in a real operation context, using real network data from a portion of the Italian 150 kV sub-transmission network.


The obtained results have supplied promising responses to the described problems, opening the way to further research in the field of data-driven future system event prediction.




\begin{thebibliography}{29}
\bibitem{voltage_predict1}
M. Hassanzadeh, C. Y. Evrenosoglu and L. Mili, ``A short-term nodal voltage phasor forecasting method using temporal and spatial correlation,'' \emph{IEEE Trans. Power Syst.}, vol. 31, n. 5, pp. 3881-3890, Sep. 2016.

\bibitem{voltage_predict2}
R. Dobbe et al., ``Linear single- and three-phase voltage forecasting and bayesian state estimation with limited sensing,'' \emph{IEEE Trans. Power Syst.}, vol. 35, no. 3, pp. 1674-1683, May 2020.

\bibitem{voltage_predict3}
A. F. Bastos, S. Santoso, V. Krishnan and Y. Zhang, ``Machine Learning-based prediction of distribution network voltage and sensor allocation,'' in \emph{Proc. 2020 IEEE Power \& Energy Society General Meeting}, pp. 1-5.

\bibitem{review_se}
K. Dehghanpour, Z. Wang, J. Wang, Y. Yuan and F. Bu, ``A survey on state estimation techniques and challenges in smart distribution systems,'' \emph{IEEE Trans. Smart Grid}, vol. 10, no. 2, pp. 2312-2322, Mar. 2019.

\bibitem{de2020review} F. De Caro, A. Andreotti, R. Araneo, M. Panella, A. Vaccaro, and D. Villacci, ``A review of the enabling methodologies for knowledge discovery from smart grids data'', \emph{Energies}, vol. 13, n. 24, pp. 1-25, 2020.


\bibitem{ml_load_forecast}
A. Almalaq and G. Edwards, ``A Review of Deep Learning Methods Applied on Load Forecasting,''in \emph{Proc. 16th IEEE International Conference on Machine Learning and Applications (ICMLA)}, 2017, pp. 511-516.

\bibitem{ml_fault}
Y. Wang, X. Wang, Y. Wu and Y. Guo, ``Power system fault classification and prediction based on a three-layer data mining structure,'' \emph{IEEE Access}, vol. 8, pp. 200897-200914, 2020.

\bibitem{ml_reliability_review}
L. Duchesne, E. Karangelos and L. Wehenkel, ``Recent developments in machine learning for energy systems reliability management," in \emph{Proceedings of the IEEE}, vol. 108, no. 9, pp. 1656-1676, Sept. 2020.

\bibitem{dt_pmu}
R. Diao et al., ``Decision tree-based online voltage security assessment using pmu measurements,'' \emph{IEEE Trans. Power Syst.}, vol. 24, no. 2, pp. 832-839, May 2009.

\bibitem{drl}
J. Duan et al., ``Deep-reinforcement-learning-based autonomous voltage control for power grid operations,'' \emph{IEEE Trans. Power Syst.}, vol. 35, no. 1, pp. 814-817, Jan. 2020.

\bibitem{dlnn}
M. Mokhtar et. al., ``Prediction of voltage distribution using deep learning and identified key smart meter locations,'' \emph{Energy and AI}, vol. 6, Dec. 2021.


\bibitem{situation_aw}
M. Panteli, P. A. Crossley, D. S. Kirschen and D. J. Sobajic, ``Assessing the impact of insufficient situation awareness on power system operation," \emph{IEEE Trans. Power Syst.}, vol. 28, no. 3, pp. 2967-2977, Aug.2013.

\bibitem{tharwat2020classification}
A. Tharwat, ``Classification assessment methods,'' \emph{Applied Computing and Informatics}, vol. 17 no. 1, pp. 168-192, 2021. 

\bibitem{bradley1997use}
A. P. Bradley, ``The use of the area under the ROC curve in the evaluation of machine learning algorithms,'' \emph{Pattern Recognition}, vol. 30, n. 7, pp. 1145-1159, Jul. 1997.

\bibitem{chicco2020advantages}
D. Chicco and G. Jurman,``The advantages of the Matthews correlation coefficient (MCC) over F1 score and accuracy in binary classification evaluation,'' \emph{BMC Genomics}, vol. 21, 2020.

\bibitem{strbac_dt}
A-A. B. Bugaje, J. L. Cremer, M. Sun and G. Strbac,``
Selecting decision trees for power system security assessment,'' \emph{Energy and AI}, vol. 6, 2021.

\bibitem{lewis2000introduction}
R. J. Lewis, ``An introduction to classification and regression tree (CART) analysis,'' presented at the 2020 Annual Meeting of the Society for Academic Emergency Medicine.

\bibitem{breiman2001random}
L. Breiman, ``Random Forests,'' \emph{Machine Learning}, vol. 45, pp. 5–32, 2001.

\bibitem{fix1989discriminatory}
E. Fix and and J. L. Hodges, ``Discriminatory Analysis-Nonparametric Discrimination: Consistency Properties,'' \emph{International Statistical Review}, vol. 57, no. 3, pp. 238–47, 1989.

\bibitem{corinna1995vapnik}
C. Cortes and V. Vapnik, ``Support-vector networks,'' \emph{Machine Learning}, vol. 20, pp. 273–297, 1995.

\bibitem{friedman1997bayesian}
N. Friedman, D. Geiger and M. Goldszmidt,``Bayesian network classifiers,'' \emph{Machine Learning}, vol. 29, pp. 131–163, 1997.

\bibitem{balakrishnama1998linear} 
S. Balakrishnama, and A. Ganapathiraju, ``Linear discriminant analysis-a brief tutorial,'' \emph{Mississippi State Univ.}, 1998, pp. 1-8.

\bibitem{carpinone2015markov} 
A. Carpinone, M. Giorgio, R. Langella, A. Testa, ``Markov chain modeling for very-short-term wind power forecasting,'' \emph{Electric Power System Research}, vol. 11, pp. 152-158, May 2015.


\end{thebibliography}
\end{document}